\documentclass[10pt,twocolumn,letterpaper]{article}

\PassOptionsToPackage{table,dvipsnames}{xcolor}

\usepackage{cvpr}              

\definecolor{cvprblue}{rgb}{0.21,0.49,0.74}
\usepackage[pagebackref,breaklinks,colorlinks,allcolors=cvprblue]{hyperref}

\usepackage{multirow}
\usepackage{amssymb}

\def\paperID{*****} 
\def\confName{CVPR}
\def\confYear{2026}

\title{Understanding-Enhanced Model Collaboration for Long-Tailed \\Egocentric Mistake Detection}

\author{
	Boyu Han\textsuperscript{1,2}\hspace{1em} Qianqian Xu\textsuperscript{1,3,}\thanks{Corresponding authors.}\hspace{1em} Shilong Bao\textsuperscript{2}\hspace{1em} Zhiyong Yang\textsuperscript{2} \\Ruochen Cui\textsuperscript{4,5} \hspace{1em}  Qingming Huang\textsuperscript{2,1,*} \\
	{\textsuperscript{1} State Key Laboratory of AI Safety, Institute of Computing Technology, CAS} \\
	{\textsuperscript{2} School of Computer Science and Tech., University of Chinese Academy of Sciences} \\
    {\textsuperscript{3} Beijing Academy of Artificial Intelligence} \\
    {\textsuperscript{4} Institute of Information Engineering, CAS} \\
    {\textsuperscript{5} School of Cyber Security, University of Chinese Academy of Sciences} \\
	{\tt\small \{hanboyu23z, xuqianqian\}@ict.ac.cn \hspace{1em} \{baoshilong,yangzhiyong21,qmhuang\}@ucas.ac.cn } \\ 
    {\tt\small cuiruochen25@mails.ucas.ac.cn}
}

\begin{document}
\maketitle
\begin{abstract}
In this report, we address the problem of determining whether a user performs an action incorrectly from egocentric video data. To this end, we propose an Understanding-Enhanced Model Collaboration Method (UE-MCM) that combines efficient coarse-grained video understanding with accurate fine-grained action reasoning. Specifically, UE-MCM contains a small model branch and a large model branch. The large model branch focuses on whether the fine-grained action itself is executed incorrectly, while the small model branch jointly takes the coarse-grained video and fine-grained segment as input to identify actions that may be locally correct but inconsistent with the overall workflow. The small model branch is built on a CLIP4CLIP video encoder initialized from a CLIP model enhanced by Diffusion Contrastive Reconstruction, and the large model branch uses the Qwen3-VL Embedding model to extract high-capacity representations from fine-grained action segments. The small-branch prediction and the large-branch prediction are then adaptively fused by a lightweight collaboration gate. To handle the long-tailed distribution of mistake instances, we optimize the classifiers with complementary objectives, including reweighted cross-entropy, AUC-oriented learning, and label-aware adjustment. The resulting system balances speed and accuracy, making it effective for detecting subtle, rare, and ambiguous mistakes in egocentric instructional videos.
\end{abstract}
    
\section{Introduction}
\label{sec:introduction}

This document introduces the solution proposed by the MR-CAS team for the Mistake Detection Challenge of the HoloAssist 2026 competition~\cite{wang2023holoassist}. The goal of the task is to determine whether a user performs an action correctly in egocentric videos. Compared with standard action recognition, mistake detection requires not only recognizing what action is being performed, but also judging whether the execution deviates from the expected procedure. This makes the task sensitive to temporal context, hand-object interactions, and subtle visual differences between correct and incorrect operations.

The problem is challenging for \textbf{two main reasons}. \textbf{First}, mistake samples are rare and often ambiguous, resulting in a long-tailed binary distribution where conventional cross-entropy training tends to overfit the dominant correct class. \textbf{Second}, mistakes can occur at different semantic levels. Some mistakes are local execution errors, where the current fine-grained action itself is wrong. Others are procedural errors, where an action may be executed correctly in isolation but is inappropriate for the current stage of the coarse-grained workflow. Using a single model~\cite{leng2024hypersdfusion,leng2023dynamic,wang2026dynamic,han2026lightfair} to cover both types of information is often inefficient and may weaken either coarse contextual understanding or fine-grained action reasoning.

\begin{figure*}
  \centering
  \includegraphics[width=\linewidth]{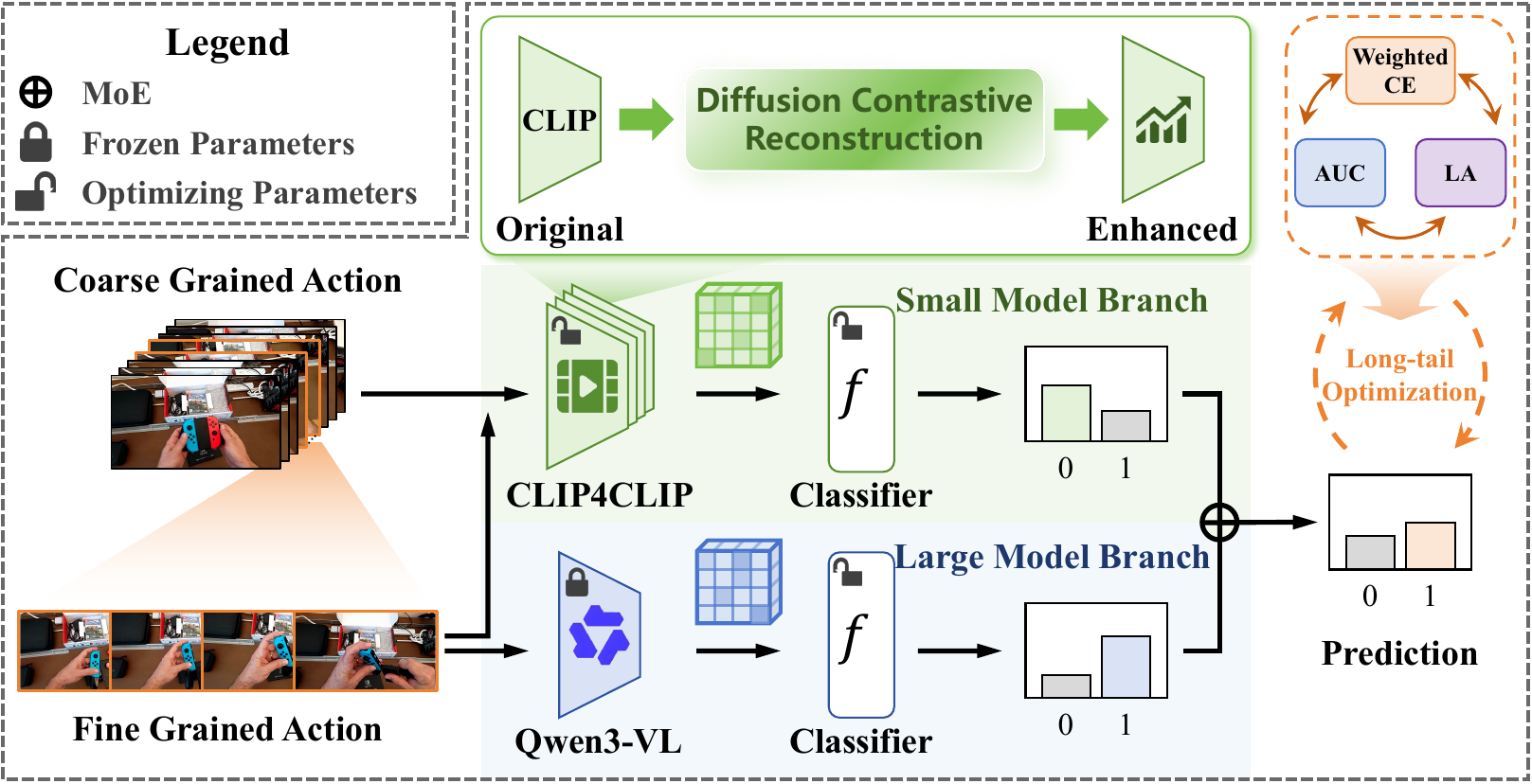}
  \caption{\textbf{An overview of our UE-MCM.} The large model branch uses Qwen3-VL Embedding to determine whether the fine-grained action itself contains a mistake. The small model branch uses a DCR-enhanced CLIP4CLIP encoder to jointly encode the coarse-grained video and the fine-grained segment, thereby reasoning about whether the action is consistent with the overall workflow.}
  \label{fig: overview}
\end{figure*}

To address these challenges, we design an \textbf{Understanding-Enhanced Model Collaboration Method (UE-MCM)}. UE-MCM contains a small model branch and a large model branch with different responsibilities. The large model branch focuses on fine-grained action correctness and predicts whether the observed action itself contains an execution mistake. The small model branch performs fast coarse-grained video understanding by jointly observing the coarse-grained video and the fine-grained segment, which helps identify cases where an action looks correct locally but is wrong in the current workflow. We first enhance the visual representation of CLIP~\cite{radford2021learning} using Diffusion Contrastive Reconstruction (DCR)~\cite{han2026dcr}. The enhanced CLIP is then used to construct a CLIP4CLIP-style video encoder~\cite{luo2022clip4clip} for the small branch. In parallel, the large model branch adopts Qwen3-VL Embedding~\cite{li2026qwen3} to extract semantically rich features from fine-grained action segments.

The predictions from the two branches are finally integrated by an adaptive collaboration gate, allowing the system to balance workflow-level consistency reasoning and action-level correctness judgment for each input. During optimization, we further combine multiple long-tail learning objectives, including reweighted cross-entropy~\cite{cui2019class}, AUC-oriented learning~\cite{yang2021learning,han2024aucseg}, and label-aware adjustment~\cite{menon2020long}. These objectives improve mistake recall, decision ranking, and probability calibration under skewed data distributions.

Overall, UE-MCM integrates action-level mistake reasoning, workflow-level consistency reasoning, representation enhancement, and long-tail optimization. The following sections describe the proposed method and the experimental settings used for the HoloAssist mistake detection benchmark.

\section{Method}
\label{sec:method}

\subsection{Overview}
\label{subsec:Overview}

Given an egocentric video, the task is to predict whether the target action contains a mistake. We denote the coarse-grained action video as $V^{c}_{0:T}$, where $T$ is the total temporal length, and the fine-grained action segment as $V^{f}_{t:t+\tau}$, where $t \geq 0$ and $t+\tau \leq T$. The model outputs a binary prediction, where label $1$ indicates a mistake and label $0$ indicates a correct action.

\Cref{fig: overview} illustrates the proposed UE-MCM. The framework contains two complementary model branches. The large model branch is a Qwen3-VL Embedding encoder that takes only the fine-grained segment $V^{f}_{t:t+\tau}$ as input and judges whether the action itself is incorrectly executed. The small model branch is a DCR-enhanced CLIP4CLIP encoder that takes both the coarse-grained video $V^{c}_{0:T}$ and the fine-grained segment $V^{f}_{t:t+\tau}$ as input, allowing it to judge whether the action is consistent with the overall workflow. The two branch predictions are fused by an adaptive collaboration gate. During training, we use long-tail optimization objectives to improve the recognition of rare mistake samples.

\subsection{Small Model Branch}
\label{subsec:small_branch}

The small model branch is designed for efficient workflow-level video understanding. CLIP provides strong image-level semantic priors through large-scale image-text contrastive pre-training~\cite{radford2021learning}. However, directly applying CLIP to egocentric mistake detection is insufficient because subtle mistakes often depend on both visual details and procedural context.

To strengthen the visual encoder, we employ Diffusion Contrastive Reconstruction (DCR)~\cite{han2026dcr}. DCR improves CLIP by introducing reconstruction-guided contrastive signals, encouraging the visual representation to preserve both class-discriminative semantics and detail-aware perceptual information. We use the enhanced CLIP visual encoder as the frame-level backbone of CLIP4CLIP~\cite{luo2022clip4clip}. The small model branch encodes both the coarse-grained video and the fine-grained segment:
\begin{equation}
    \mathbf{h}^{c}_{s} = \phi_{s}\left(V^{c}_{0:T}\right), \quad
    \mathbf{h}^{f}_{s} = \phi_{s}\left(V^{f}_{t:t+\tau}\right),
\end{equation}
where $\phi_s$ denotes the DCR-enhanced CLIP4CLIP encoder. The two representations are fused before classification:
\begin{equation}
    \mathbf{h}_{s} =
    \psi_s\left([\mathbf{h}^{c}_{s}; \mathbf{h}^{f}_{s}; \mathbf{h}^{c}_{s}\odot \mathbf{h}^{f}_{s}]\right),
\end{equation}
where $\psi_s(\cdot)$ is a lightweight fusion projection, and $\odot$ denotes element-wise multiplication. Since this branch is lightweight and observes both temporal scopes, it provides stable coarse action context and helps detect cases where an action may be correct in isolation but wrong within the current procedure.

\subsection{Large Model Branch}
\label{subsec:large_branch}

The large model branch focuses on fine-grained action-level mistake reasoning. We use Qwen3-VL Embedding~\cite{li2026qwen3} to encode the fine action segment. Compared with the small model branch, Qwen3-VL has stronger multimodal representation capacity and is better suited for recognizing whether the observed manipulation itself is incorrectly executed. For each fine-grained segment, the large model branch extracts
\begin{equation}
    \mathbf{h}_{l} = \phi_{l}\left(V^{f}_{t:t+\tau}\right),
\end{equation}
where $\phi_l$ denotes the Qwen3-VL Embedding model. We freeze the large embedding model during classifier training and use its output for the large-branch classifier. This keeps optimization efficient while preserving the semantic strength of the large model.

\subsection{Model Collaboration}
\label{subsec:model_collaboration}

The two branches provide complementary information. The large model branch predicts action-level mistakes from fine-grained execution cues, while the small model branch predicts workflow-level inconsistencies by jointly observing the coarse-grained video and fine-grained segment. The branch representations are fed into their corresponding classification heads:
\begin{equation}
    \mathbf{z}_{s} = g_s(\mathbf{h}_{s}), \quad
    \mathbf{z}_{l} = g_l(\mathbf{h}_{l}),
\end{equation}
where $g_s$ and $g_l$ are lightweight classification heads, and $\mathbf{z}_s,\mathbf{z}_l \in \mathbb{R}^{2}$ are logits for correct and mistake classes.

To adaptively combine the two predictions, we use a collaboration gate. The gate takes the projected branch features as input and outputs normalized branch weights:
\begin{equation}
    \bar{\mathbf{h}}_{s} = P_s(\mathbf{h}_{s}), \quad
    \bar{\mathbf{h}}_{l} = P_l(\mathbf{h}_{l}),
\end{equation}
\begin{equation}
    [\alpha_s, \alpha_l] =
    \operatorname{softmax}\left(W_g[\bar{\mathbf{h}}_{s};\bar{\mathbf{h}}_{l};\bar{\mathbf{h}}_{s}\odot \bar{\mathbf{h}}_{l}] + \mathbf{b}_g\right),
\end{equation}
where $P_s(\cdot)$ and $P_l(\cdot)$ project branch features into a shared fusion space, and $\odot$ denotes element-wise multiplication. The final logits are computed as
\begin{equation}
    \mathbf{z} = \alpha_s \mathbf{z}_{s} + \alpha_l \mathbf{z}_{l}.
\end{equation}
This design keeps the branch inputs explicit: the small branch combines coarse and fine videos for workflow-level reasoning, while the large branch focuses on the fine segment for action-level reasoning. Prediction-level collaboration then dynamically balances the two decisions.

\subsection{Long-Tail Optimization}
\label{subsec:long_tail_optimization}

Mistake detection is naturally imbalanced because correct actions appear much more frequently than mistakes. We therefore optimize the classifiers with complementary long-tail objectives.

\noindent \textbf{Reweighted CE Loss~\cite{cui2019class}.} We use class-rebalanced cross-entropy to increase the penalty for underrepresented mistake samples:
\begin{equation}
    \mathcal{L}_{WCE} = - \sum_{i=1}^{N} w_{y_i}\log p_{i,y_i},
\end{equation}
where $p_{i,y_i}$ is the predicted probability of the ground-truth class and $w_{y_i}$ is computed according to class frequency.

\noindent \textbf{AUC Loss~\cite{yang2021learning}.} To improve ranking quality under class imbalance, we adopt an AUC-oriented objective:
\begin{equation}
    \mathcal{L}_{AUC} =
    \frac{1}{n^+n^-}
    \sum_{i=1}^{n^+}\sum_{j=1}^{n^-}
    \ell\left(s_j^- - s_i^+\right),
\end{equation}
where $s_i^+$ and $s_j^-$ are mistake and correct scores, respectively. This loss encourages mistake samples to receive higher scores than correct samples.

\begin{table*}[t!]
  \centering
  \renewcommand\arraystretch{1}
  \caption{\textbf{Results obtained on the test set.} The champion and the runner-up are highlighted in \textbf{bold} and \underline{underline}.}
  \resizebox{0.8\linewidth}{!}{
    \begin{tabular}{ccccccc}
    \toprule
    \multirow{2}[2]{*}{\textbf{Method}} & \multirow{2}[2]{*}{\textbf{Modality}} & \multirow{2}[2]{*}{\textbf{F-score}} & \multicolumn{2}{c}{\textbf{Correct}} & \multicolumn{2}{c}{\textbf{Mistake}} \\
          &       &       & \textbf{Precision} & \textbf{Recall} & \textbf{Precision} & \textbf{Recall} \\
    \midrule
    Random & -     & 0.28  & 0.61  & 0.10  & \textbf{0.15} & 0.46  \\
    \midrule
    \multicolumn{1}{c}{\multirow{4}[2]{*}{TimeSformer~\cite{wang2023holoassist} \textsubscript{\textcolor[rgb]{ .933,  .51,  .184}{(Baseline)}}}} & RGB   & 0.35  & 0.83  & 0.52  & \underline{0.13}  & 0.27  \\
          & Hands & 0.40  & 0.93  & 0.52  & \underline{0.13}  & 0.31  \\
          & RGB+Hands & 0.36  & 0.86  & 0.43  & 0.10  & 0.12  \\
          & RGB+Hands+Eyes & 0.32  & 0.89  & 0.43  & 0.11  & 0.50  \\
    \midrule
    \multicolumn{1}{c}{UNICT Solution \textsubscript{\textcolor[rgb]{ .933,  .51,  .184}{(2024 Top1)}}} & RGB+Eyes & 0.51  & \underline{0.95}  & \textbf{0.93} & 0.06  & 0.09  \\
    \midrule
    \multicolumn{1}{c}{MR-CAS Solution~\cite{han2025dual} \textsubscript{\textcolor[rgb]{ .933,  .51,  .184}{(2025 Top1)}}}  & RGB   & \underline{0.57} & \textbf{0.97} & 0.60  & 0.08  & \textbf{0.63} \\
    \midrule
    \rowcolor[rgb]{ .992,  .91,  .855} Ours  & RGB   & \textbf{0.60} & \textbf{0.97} & \underline{0.72}  & 0.11  & \underline{0.62} \\
    \bottomrule
    \end{tabular}%
    }
  \label{tab: results}%
\end{table*}%

\noindent \textbf{Label-Aware Loss~\cite{menon2020long}.} We further use label-aware adjustment to calibrate the decision boundary for long-tailed data:
\begin{equation}
    \mathcal{L}_{LA} =
    \sum_{i=1}^{N}
    \ell\left(\mathbf{z}_{i} + \log \boldsymbol{\pi}, y_i\right),
\end{equation}
where $\boldsymbol{\pi}$ denotes the empirical class prior. The final objective is
\begin{equation}
    \mathcal{L} =
    \mathcal{L}_{WCE}
    + \lambda_{AUC}\mathcal{L}_{AUC}
    + \lambda_{LA}\mathcal{L}_{LA}.
\end{equation}
By combining these objectives, the classifier learns from both calibrated class priors and pairwise ranking constraints, improving robustness for rare mistakes.

\section{Experiments}
\label{sec:experiments}

In this section, we describe some details of the experiments and present our results.

\subsection{Implementation Details}
\label{subsec:Implementation Details}

We conduct all experiments using eight NVIDIA A100 GPUs. Frames are uniformly sampled from both the full coarse-grained video and the fine-grained segment. The backbone encoders are frozen during classifier training, and only the projection layers, classification heads, and collaboration gate are optimized. Branch features are projected into a shared hidden space before collaborative fusion. The classifier heads are implemented as lightweight multilayer perceptrons with dropout. We train the system with AdamW and a cosine learning-rate schedule. The entire model is optimized using the Adam optimizer with a learning rate of $1 \times 10^{-5}$. We set the batch size to $128$ clips, each consisting of $32$ frames. The model is trained for a total of $5$ epochs.

\subsection{Results}
\label{subsec:Results}

\Cref{tab: results} presents the performance of various models on the mistake detection task. Compared to Random and TimeSformer, our method significantly improves the F-score. Furthermore, in comparison to the top-performing method of 2024, our method achieves a substantial improvement in mistake recall. Compared with the top-performing method of 2025, our method further improves correct recall. Most notably, our method attains competitive performance using only the RGB modality, matching or even surpassing models that rely on multimodal inputs.
\newpage
{
    \small
    \bibliographystyle{ieeenat_fullname}
    \bibliography{main}
}


\end{document}